**FLYING, HOPPING PIT-BOTS FOR CAVE AND LAVA TUBE EXPLORATION ON THE MOON AND MARS.** J. Thangavelautham, M. S. Robinson, A. Taits, T. J. McKinney, S. Amidan, A. Polak, School of Earth and Space Exploration, Arizona State University, 781 E. Terrace Mall, ISTB4-795, jekan@asu.edu

**1. Introduction:** Wheeled ground robots are limited from exploring extreme environments such as caves, lava tubes and skylights. Small robots that utilize unconventional mobility through hopping, flying and rolling can overcome many roughness limitations and thus extend exploration sites of interest on Moon and Mars. In this paper we introduce a network of 3 kg, 0.30 m diameter ball robots (pit-bots) that can fly, hop and roll using an onboard miniature propulsion system (**Fig. 1**). These pit-bots can be deployed from a lander or large rover. Each robot is equipped with a smartphone sized computer, stereo camera and laser rangefinder to perform navigation and mapping. The ball robot can carry a payload of 1 kg or perform sample return. Our studies show a range of 5 km and 0.7 hours flight time on the Moon.

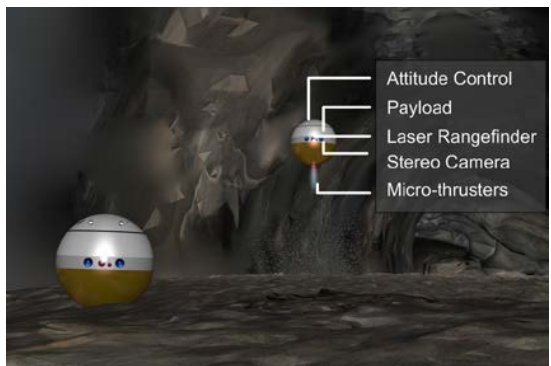

*Fig. 1 – Pit-bot Cave Explorer Concept*

**2. Extreme Environment Exploration:** High resolution orbital imagery from LROC revealed evidence for subsurface voids and mare-pits on the lunar surface [1, 2]. Mare Ingenii shown in **Fig. 2** is 70 m deep and is theorized to be collapsed entrance to a lava tube. The rugged terrain inside a lava tube entrance, with slopes steeper than 30° make it impassable by conventional wheeled robots. Accessible voids could be used for a future human base because they offer a natural radiation and micrometeorite shield and offer constant habitable temperatures of -20 to -30 °C [11].

Hopping bots [3-5] using mechanical systems are insufficient because of the expected rugged environment particularly, when the slopes are too steep. Hopping poses challenges in determining where to land gently, particularly in rugged environment. In contrast flying allows for the systems to gently take off and land at a desired landing spot minimizing impact forces. Other methods such as tethering a probe to the base rover will not work in caves and lava tubes, because these formations are not straight, instead they are known to zig-zag. In addition tethers can catch on sharp rocks, displace rocks and risk tangling both the bot and the base rover. In contrast a flying robot is physically untethered to the rover and any risks it experiences leaves the rover unaffected.

Current technology is severely limited by energy density of batteries and from miniature propulsion systems [6]. These power constraints constrain mission duration, mobility and overall functionality of the small probes. To overcome the power problem we leverage advancements in miniaturized chemical mobility systems together with integrating the required navigation and autonomous control technology into a small ball-shaped probe.

**3. Flying and Hopping Pit-bots:** Our proposed design consists of a network of 3 or more pit-bots (**Fig. 3**) to perform extreme environment exploration. The lower half of the sphere contains the propulsion system, with storage tanks for RP1 and hydrogen peroxide.

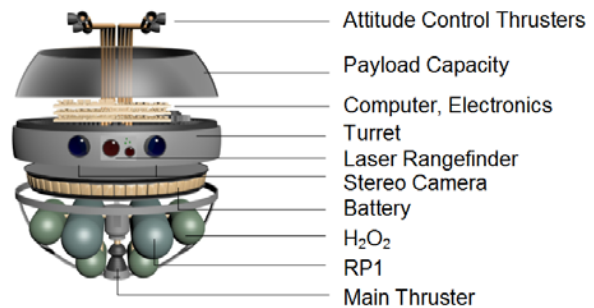

*Fig. 3 – Pit-bot Internals*

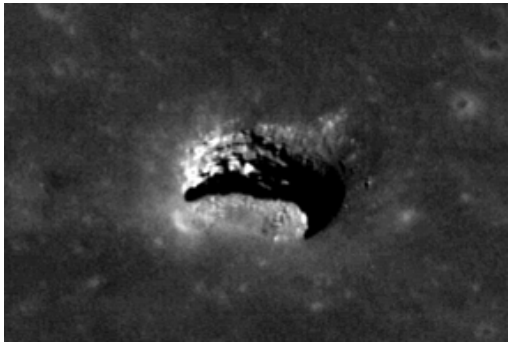

*Fig. 2 – Mare Ingenii*

The attitude control system is in the top and contains micro-thrusters for maintaining yaw, pitch and roll.

Next is the Lithium Thionyl Chloride batteries arranged in a circle as shown. The mass budget is shown in **Table 1**. Comparison with other mobility options, including use of Radioisotope Thermal Generators (RTGs) and batteries show our design was found to be the only one to meet a minimal set of requirements (**Table 2**). A pair of stereo cameras and a laser range finder rolls on a turret enabling the pit-bot to take panoramic pictures and scan the environment without having to move using the propulsion system. Above the turret are two computer boards, IMU and IO-expansion boards and a power board.

*Table 1: Pit Bot Mass Budget*

| Major Subsystem | Mass (kg) |
|---|---|
| Propulsion | 1.2 |
| Computer, Comms, Electronics | 0.2 |
| Power | 0.3 |
| Stereo Camera, Laser Ranger | 0.3 |
| Payload | 1.2 |
| Total | 3 |

*Table 2: Technology Comparison*

| Power & Propulsion Technology | Specific Energy Wh/kg | Mass [kg] | Waste Heat [W] | Fly Time [hr] | Range [km] |
|---|---|---|---|---|---|
| Proposed System $RP_1 + H_2O_2$, $LiSOCl_2$ | 700 | 1.5 | 5 | 0.7 | 5 |
| $LiSOCl_2$ | 700 | 3 | 0.5 | 0 | 0.5 |
| RTG | $\gg 10^6$ | 4.0 | 100 | 0 | 0.4* |
| Lithium Ion | 130 | 10 | 0.5 | 0 | <0.1* |

**4. Pit-Bot Propulsion:** The critical subsystem required for this pit-bot is the propulsion system. The robots shall contain one primary lift engine positioned at the vehicles bottom portion and 8 "warm-gas" attitude control thrusters positioned at the top of the bot. For the ball robot propulsion we hereby consider RP1-$H_2O_2$ engine. Hydrogen-Peroxide is the oxidizer as well as the propellant for the Attitude Control System (ACS). Other oxidizers were considered for this robot including water, liquid-oxygen, and liquid nitrous-oxide. However, for the application of these small pit-bots, these oxidizers will not work. To begin, water may only be used to oxidize metal-hydrides and is not practical for use in an ACS since no source of heat is available to generate the required quantities of vapor. Liquid-oxygen requires cryogenic storage that is impractical due to the size constraints of a 30 cm diameter vehicle within the lunar caves.

Liquid nitrous-oxide requires immense pressures (7 MPa) for liquid storage and is quite difficult to accomplish from a safety stand-point. Hydrogen peroxide is a good option because it can be tested at first with low purities (dissolved in water) to validate our physical models and predictions. This minimizes risks during system development. Successful implementation at low purities will give us the confidence to increase to 50 % concentration.

For a non-cryogenic fuel, RP-1 has by far the highest storage density of approximately 700kg/m$^3$. Furthermore, RP-1 is relatively low-cost, non-toxic, and easy to handle[7]. RP-1/$H_2O_2$ thrusters have been used since the 1960's by the Soviet Union and have achieved TRL-9. However our efforts will be focused in miniaturizing the RP-1 $H_2O_2$ engine for the ball robot system (see Fig. 4). To implement this system in a small volume and avoid the use of pumps and mechanical devices, our design uses pressurized nitrogen gas to initiate transport of the reactants into the combustion chamber. Prior to being injected into the main rocket-engine or the ACS valves, the hydrogen-peroxide is decomposed by means of a silver catalyst into oxygen and water. In the process of catalyzed decomposition, the oxygen and water will heat-up to a temperature of 600 $^o$C. When the warm oxygen/water (oxidizer) is used to power the ACS system, the resulting specific impulse is approximately 180 seconds (no combustion). It is predicted with this engine design, a specific impulse of 330 seconds will be achieved 50 % $H_2O_2$ concentration.

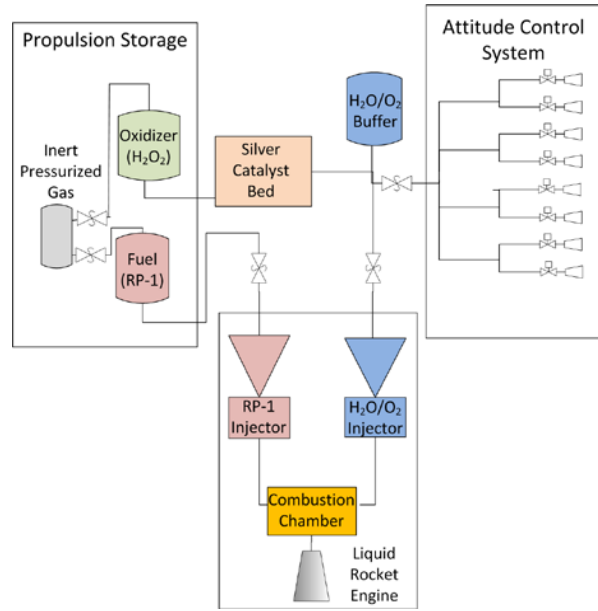

*Fig. 4 – Pit-bot Propulsion System*

**5. Pit-bot Navigation and Mapping:** The pit-bot would navigate by autonomously forming a triangular formation (**Fig. 5, 6**). The robots are equipped with bright lights that serve as beacon or as light sources in the lava tube/cave. Each robot moves forward, one robot at a time a short distance much like a bucket brigade [8-10]. Each robot takes stereo ground images, just before descending to the ground with one or both of the other robots in view. Because the ground robots have bright lights, a simple blob detection algorithm is sufficient to locate the ball robots in an image. Converting the stereo image to point cloud, provides distance estimates to the ball robots on the ground. The robots will have sufficient computational capabilities to process stereo images. Using these distances, it is possible to estimate the position of the ball robot relative to other robots on the ground (**Fig. 7**).

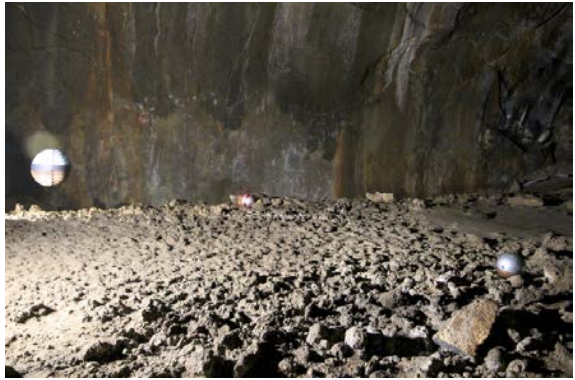

*Fig. 5- A network of 3 ball robots in a lava tube (1 flying mockup consists of a quad copter). Two on the ground are static display. Our studies show that the flying robot can locate/identify other robots within a 7 m distance.*

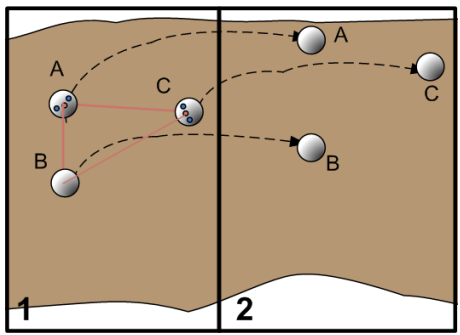

*Fig. 6- The ball robots maintain a triangular formation, as each in order A, B and C take a short flight or hop to its next resting stop. Once they are in the designated triangle formation (inset 1), then lasers will be used to triangulate distance.*

If the other robots are visible, when all three robots are on the ground, the laser range finder could be used to get even more accurate distance measurements through direct triangulation. These positions would be recorded giving a total estimate of the position travelled by each robot from the base rover. Sections of a lava tube could be mapped (**Fig. 8**). Commercial point laser rangefinder such as from Leica Disto E7100i have an error of 0.0025 % with a maximum range of 70 m. Using these estimates, the robots would be taking measurements every 9 to 5 meters interval. We would expect the total error in positioning using our approach to be 0.3 % to 0.5 % for 1 km radial distance.

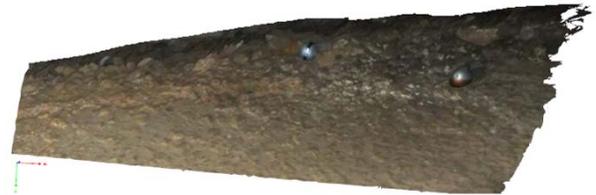

*Fig. 7- Stereo images taken to produce 3D point cloud and mesh images of the pit-bots inside a lava tube (Flagstaff, Arizona).*

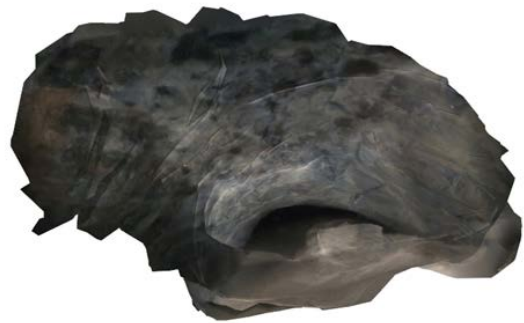

*Fig. 8 - The pit-bots will obtain 3D images and produce 3D maps of interiors as demonstrated in Government cave (lava tube) near Flagstaff, Arizona.*

**6. Pit-bot Operations and Control:** The pit-bots are intended to be fully autonomous. They will have the ability to hop, fly, hover, and roll. The robots will most often perform a fly-hop, which provides all the advantages of hop, but with a soft landing. Optimal fuel saving trajectories have been found to obtain maximum hop range for given rocket engine specific impulse (**Fig. 9**). In addition, one of the goals of the pit-bot propulsion and attitude control system is to achieve hovering capability equivalent to current quad-copters. This hovering mode will be used to build 3D panaromic maps and for tracking the other pit-bots. A mission planner specifies the target coordinates where the ball robots and the payload package are delivered. At this point the pit-bots develop an internal navigation path avoiding obstacles in the path. In this approach each ball robot operates cooperatively, without a centralized supervisor or leader to mitigate damage from loss of one or more robots.

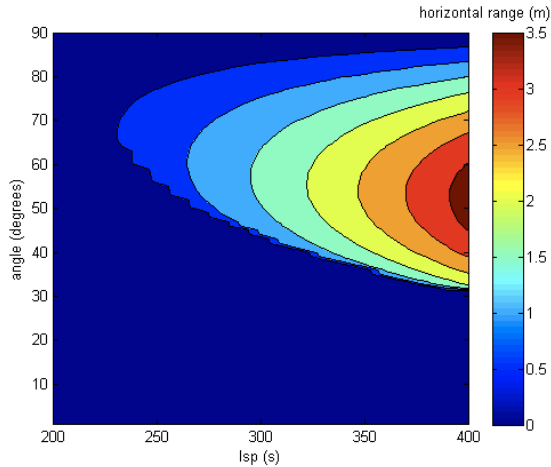

*Fig. 9 – Comparison of pit-bot fly-hop trajectories to minimize fuel while maximizing range.*

**7. Conclusions and Future Work:** Detailed concept studies of pit-bot design and use are ongoing. The results to date show the principal feasibility of the navigation and controls approach. Development of an attitude control system also shows promising results. However, significant challenges remain in the development of the propulsion system even though the propulsion technology is mature. The challenge will be in integration and miniaturization of the system into a 30-cm sphere.